\documentclass[11pt]{article}

\usepackage[final]{acl}

\usepackage{times}
\usepackage{latexsym}
\usepackage[fleqn]{amsmath}
\usepackage{amsfonts}

\usepackage[T1]{fontenc}

\usepackage[utf8]{inputenc}

\usepackage{microtype}

\usepackage{inconsolata}

\usepackage{graphicx}
\usepackage{booktabs}

\title{Relational Attention for Data-Efficient Language Modeling}

\author{Adrian Brasoveanu \\
  UC Santa Cruz, \\
  \texttt{abrsvn@gmail.com} \\\And
  Ece Takmaz \\
  Utrecht University \\
  \texttt{e.k.takmaz@uu.nl} \\\And
  Jakub Dotla\v{c}il \\
  Utrecht University \\
  \texttt{j.dotlacil@uu.nl} \\}

\begin{document}
\maketitle
\begin{abstract}
    We present Relational BabyLM, a system submission to the BabyLM 2026 challenge that combines two cognitively motivated inductive biases in a single decoder-only Transformer. Architecturally, we replace standard self-attention with a Dual Attention Transformer (DAT), which separates the routing of object-level (``sensory'') lexical features from structural/relational information \citep{altabaa2025disentangling,altabaa2024abstractors,webb2024relational,kerg2022neural,esbn}. Relational attention (RA) disentangled from self-attention greatly increases data efficiency and out-of-training-sample generalization on purely relational tasks, but language modeling requires object-level and relational information to be integrated as well as disentangled, and RA-based LMs have remained largely unexplored. BabyLM's data-constrained training and comprehensive evaluation is an ideal testing ground for whether that data efficiency transfers. As a training intervention, we add a Next-Latent Prediction (NextLat; \citealt{teoh2026nextlatent}) objective that encourages hidden states to compress history incrementally into a dense belief state. Architecture is the dominant factor for structural linguistic generalization; the objective is secondary but still significant. DAT's three relational attention types (full RA vs.\ the simpler RCA and DisRCA variants) are largely interchangeable at 10M words; full RA pulls ahead at 100M. We also introduce a novel symbol-retrieval mechanism (RoPE-based, as opposed to learned, relative symbols) that matches learned symbol libraries while adding no parameters. On the strict (100M-word) track, our best model ranks 6th of 55 overall and 3rd of 55 on the leaderboard's NLP-task subset at the time of writing; our two strongest models outperform the GPT-2 baseline on most benchmarks, with one attaining the highest EWoK score among strict-track entries.
\end{abstract}

\section{Introduction}
\label{intro}

Standard Transformer language models \citep{vaswani,radford2019language} face two structural limitations in the low-data BabyLM regime. First, ordinary self-attention entangles relational information with high-dimensional object-level features: the query--key dot product is derived from the inputs / sources / objects, and can be understood as encoding the relation between the objects and controlling which source / object is selected (via the softmax) based on that relational information, but the routed value is a feature vector derived from that same source / object. This entanglement is at odds with the division of labor in, for example, formal semantics between lexical semantics, which makes word-specific, object-like contributions to sentential meaning (truth conditions), and compositional semantics, which makes structure-based, relation-like contributions to meaning \citep{montague1973proper,dowty1981introduction,partee1995lexical}.

The ability to separate relational structures from sensory features is evident in infants as young as seven months, who can readily generalize algebraic rules (relational patterns: ABA vs.\ ABB) to novel vocabularies (objects) \citep{marcus1999rule}. Infants generalize this form of relational (symbolic) reasoning quickly, from few `training' examples, and apply it systematically to novel out-of-training-sample (OOTS) objects. In contrast, self-attention-based transformers require a lot more data, and still struggle with OOTS generalization \citep{webb2024relational,kerg2022neural,esbn}.

A second issue with self-attention is that, unlike RNNs, it allows a model to look back at any past token, so the architecture lacks an inherent pressure to compress sequence history into a dense, unified summary that can support OOTS generalization. This contrasts with human language processing, which is subject to a ``Now-or-Never'' bottleneck \citep{christiansen2016now}: linguistic input must be rapidly chunked and compressed into hierarchical representations before fleeting sensory memory decays. This forces eager incremental processing that drives expectations in rational parsing \citep{hale2011what} and cue-based memory retrieval \citep{lewis2005activation}.

In this paper, we present Relational BabyLM, a system submission to the BabyLM 2026 challenge \citep{choshen2026babylmturns4papers} that combines two cognitively motivated inductive biases in a single decoder-only Transformer. The primary architectural contribution is (our own reimplementation of) the Dual Attention Transformer (DAT; \citealp{altabaa2025disentangling}), which augments sensory self-attention with a parallel \emph{relational attention} mechanism: source selection works as usual, but what is retrieved is an explicit pairwise relation vector together with an abstract symbol identifying the source (Section~\ref{dat:attn}). DAT's data-efficiency advantages have been largely established on purely relational tasks, and the evidence for its usefulness in language modeling is limited to modest perplexity gains \citep{altabaa2025disentangling}. Whether relational attention helps a language model under a broad linguistic evaluation has remained an open question investigated by our present submission.

The complementary training intervention is Next-Latent Prediction (NextLat; \citealp{teoh2026nextlatent}), which trains a small auxiliary dynamics model to predict the next hidden state from the current one and the actual next token (RNN-style), pushing the hidden states toward a \emph{belief state}, a sufficient statistic of the past for predicting the future similar to dynamic semantics info states \citep{kamp1981theory,heim1982semantics,groenendijk1991dynamic}. The dynamics model is discarded after training, leaving the base architecture and autoregressive inference unmodified (Section~\ref{dat:nextlat}).

We submit models to the strict (100M-word) and strict-small (10M-word) tracks, and run a controlled experimental matrix on the strict-small track to separate the two inductive biases. The evaluation reveals an asymmetry between them.
\begin{enumerate}
  \setlength{\itemsep}{0pt}
  \item \textbf{Architecture is the dominant factor for structural generalization} (Sections~\ref{res:arch} and~\ref{res:sara}). DAT configurations outperform matched standard Transformers on BLiMP, with the largest gains on structural domains such as island effects and subject--verb agreement, though a head-ratio sweep shows that enlarging the relational stream, rather than merely having one, does not help further.
  \item \textbf{The training objective is a secondary but significant factor for cognitive alignment} (Section~\ref{res:nextlat}). NextLat models explain more variance in human reading measures than next-token-prediction models, and improve 5 of 7 fine-tuned (Super)GLUE task accuracies.
  \item \textbf{The symbol-retrieval mechanism matters} (Section~\ref{res:rca}). In a fully powered five-seed comparison across seven mechanisms, relational-symbolic variants underperform and the rest are indistinguishable, so our new RoPE-based relative symbols match a learned relative symbol library at no parameter cost.
\end{enumerate}

On the strict (100M-word) track, our two strongest models rank among the top ten entries on the official leaderboard at the time of writing, with our best model 6th of 55 overall and 3rd of 55 on the NLP-task subset, and both outperform the official GPT-2 baseline on most benchmarks; on the strict-small (10M-word) track, our NextLat submission ranks 8th of 127 on BLiMP and 6th on EWoK (Section~\ref{res:strict}).

\section{Model}
\label{dat}

The Dual Attention Transformer (DAT) \citep{altabaa2025disentangling} separates two kinds of information that are entangled in ordinary self-attention. Standard self-attention routes \emph{sensory information}: it selects source objects by comparing queries and keys, then routes their feature vectors. DAT adds \emph{relational attention}, whose routed values explicitly encode relations between a receiving object and source objects. The resulting layer contains both sensory heads and relational heads, allowing the model to retrieve object-level features and relation-level information in parallel.

\subsection{Sensory and Relational Attention}
\label{dat:attn}

Let $\mathbf{x}=(x_1,\ldots,x_n) \in \mathbb{R}^{n \times d}$ be a sequence of object representations, where $n$ is the sequence length and $d$ is the model (embedding) dimension, so that each $x_i \in \mathbb{R}^{d}$. Standard sensory attention from receiver $x_i$ to context $\mathbf{x}$ is
\begin{align}
\mathrm{Attn}(x_i,\mathbf{x}) &= \textstyle\sum_{j=1}^{n} \alpha_{ij}\, x_j W_v, \\
\alpha_i &= \mathrm{Softmax}\big(\langle x_i W_q,\, x_j W_k \rangle\big)_{j=1}^{n},
\end{align}
where $W_q$, $W_k$, $W_v$ are learned query, key, and value projection matrices and $\alpha_{ij}$ is the $j$-th component of $\alpha_i$. The attention weights $\alpha_{ij}$ encode a selection criterion, but the retrieved values $x_j W_v$ remain sensory representations of source objects. Relational attention keeps the selection mechanism but changes what is retrieved. Instead of routing only $x_j W_v$, the receiver retrieves a relation vector between $x_i$ and $x_j$, tagged by a symbol $s_{ij}$ identifying the source:
\begin{equation}
\begin{split}
\mathrm{RelAttn}&(x_i,\mathbf{x}) = \\
&\sum_{j=1}^{n} \alpha_{ij}\big(r(x_i,x_j)\, W_r + s_{ij}\, W_s\big),
\end{split}\label{eqrel}
\end{equation}
where the pairwise relation vector $r(x_i,x_j)\in\mathbb{R}^{d_r}$ stacks $d_r$ learned comparison channels,
\begin{equation}
r(x_i,x_j)=\sigma_{\mathrm{rel}}\!\Big(\big(\langle x_i W_q^{\mathrm{rel},\ell},\, x_j W_k^{\mathrm{rel},\ell}\rangle\big)_{\ell=1}^{d_r}\Big),
\end{equation}
each channel $\ell$ using its own relational projections $W_q^{\mathrm{rel},\ell}, W_k^{\mathrm{rel},\ell}$, and $\sigma_{\mathrm{rel}}$ being a relation activation function (e.g., identity, sigmoid, tanh, or softmax). The symbol $s_{ij}\in\mathbb{R}^{d_s}$ identifies the source as an abstract symbol (Section~\ref{dat:symbols}). The projections $W_r\in\mathbb{R}^{d_r\times d_v}$ and $W_s\in\mathbb{R}^{d_s\times d_v}$ map the relation and symbol into a common output space $\mathbb{R}^{d_v}$, so their contributions can be summed. Thus, the ``message'' from source $j$ to receiver $i$ contains a source-receiver relation $r(x_i,x_j)$ and a symbol $s_{ij}$ tagging the sender.

In principle, Transformers with only self-attention or only relational attention could be used for language modeling. The Dual Attention Transformer combines the two. For $n_h^{\mathrm{sa}}$ sensory heads and $n_h^{\mathrm{ra}}$ relational heads, the combined output for position $i$ concatenates the two groups and applies an output projection. The SA/RA head split controls the allocation to sensory versus relational retrieval: a 6SA/6RA split gives half to each stream, while 9SA/3RA gives three quarters to sensory attention. We instantiate DAT with pre-normalized causal decoder blocks, applying the same decoder mask to sensory and relational heads before the softmax over source positions.

\subsection{Symbol Assignment}
\label{dat:symbols}

Relational attention tags each source $j$ with a symbol that is receiver-conditioned, written $s_{ij}$. Some of the symbol ``libraries'' and associated retrieval mechanisms can depend on the source alone ($s_{ij}=s_j$), as with a symbol tied to the absolute position $j$, whereas other symbol libraries make it depend on both, for example, relative position symbols for the receiver--source offset $j-i$. The original DAT design \citep{altabaa2025disentangling} offers three mechanisms---symbols indexed by absolute position, symbols indexed by the receiver--source offset, and a learned symbol library retrieved via symbolic attention (or a relational-symbolic variant)---and uses symbolic attention for its language models. We add a fourth, RoPE-based relative symbols: in the original work \citep{su2023roformer} rotary encodings act on the queries and keys as a positional encoding, whereas here they generate the relative symbol (the ``value'') itself. The symbol mechanisms we study in a controlled five-seed comparison are:
\begin{itemize}
    \setlength{\itemsep}{0pt}
    \item \textbf{Learned relative symbols} (\texttt{relative}): a learned symbol library indexed by the relative offset $j-i$, clipped to a maximum distance $\Delta$. Each source is tagged by a receiver-conditioned relative symbol.
    \item \textbf{RoPE-based relative symbols} (\texttt{relative\_rope}, new): instead of a learned offset-indexed library, the relative symbol at offset $j-i$ is generated by applying rotary position encoding (RoPE) to the clipped relative distance. This gives a deterministic relative symbol that does not require learning a separate symbol library.
    \item \textbf{Learned positional symbols} (\texttt{positional}): a learned symbol library indexed by absolute source position $j$, so the symbol $s_j$ depends only on the source position in the sequence.
    \item \textbf{Sinusoidal positional symbols} (\texttt{positional\_sinusoidal}): the same as learned positional symbols, but the position-indexed symbol library is replaced by a fixed sinusoidal positional encoding, as in one of the Abstractor experiments \citep{altabaa2024abstractors}.
    \item \textbf{Symbolic attention} (\texttt{symbolic}): maps each object to a convex combination of abstract symbols from a learned library $\mathcal{S}=(s_1,\ldots,s_{n_s})$ via learned feature templates.
    \item \textbf{Relational-symbolic} (\texttt{relsymbolic(\_n4)}): first computes a local relation profile for each position over a causal neighborhood of size $k$ (default 2, or 4), then applies symbolic attention to that profile.
\end{itemize}

\subsection{Rel.\ Attn.\ Type: RA, DisRCA, and RCA}
\label{dat:mechanisms}

We compare three relational attention types, which differ in (i) whether the attention weight itself serves as the relation, and (ii) what is routed.
\textbf{RCA} \citep{altabaa2024abstractors} is self-attention where the values are replaced by symbols: $x_i' = \sum_j \alpha_{ij}\, s_{ij}$, with $\alpha_{ij} = \sigma_{\mathrm{rel}}(\langle x_i W_q, x_j W_k\rangle)$. The attention weight itself is the relation, and the routed values are symbols rather than sensory features. The Abstractor treats $\sigma_{\mathrm{rel}}$ as a configurable hyperparameter, softmax in most of its experiments but element-wise activations in others, since softmax can mask relevant information. Our runs use identity (architecture comparison, head-ratio sweep, submitted models) or sigmoid (symbol-retrieval comparison; Appendix~\ref{app:configs}).
\textbf{DisRCA} \citep{altabaa2025disentangling} is an intermediate design between RCA and RA. A softmax attention weight $\alpha_{ij}$ selects sources, which are gated by a separately computed per-head relational compatibility $r_{ij}$---the construction of Eq.~\ref{eqrel} taken as one scalar channel per relational head rather than as a $d_r$-dimensional vector: $x_i' = \sum_j \alpha_{ij}\, r_{ij}\, s_{ij}$. The elementwise product creates an AND-like condition---a source is routed only if it is both selected and relationally compatible---and only the symbol is routed. Full
\textbf{RA} (\citealt{altabaa2025disentangling}; Section~\ref{dat:attn} above) separates selection from relation. Two independent sets of query/key maps compute (i)~a softmax attention weight $\alpha_{ij}$ for selecting sources and (ii)~a separate relation vector $r(x_i,x_j)$. The output routes both: $x_i' = \sum_j \alpha_{ij}\big(r(x_i,x_j)\,W_r + s_{ij}\,W_s\big)$.

\subsection{Next-Latent Prediction and Belief States}
\label{dat:nextlat}

Self-attention's unrestricted look-back gives a Transformer no inherent pressure to compress sequence history into a dense, unified summary like RNNs do. To add this pressure with the goal of OOTS generalization, we augment the training objective with Next-Latent Prediction (NextLat; \citealp{teoh2026nextlatent}). Let $h_t$ be the final-layer hidden state at position $t$ and $p_\theta(\cdot \mid h_t)$ the LM head that maps it to a distribution over the vocabulary. A small latent dynamics model predicts the next hidden state from the current one and the next token, $\hat{h}_{t+1} = h_t + \delta_\psi([x_{t+1}; h_t])$, where the update $\delta_\psi$ is a LayerNorm over $[x_{t+1}; h_t]$ followed by a three-layer GELU MLP of width $2d$; the skip in the equation is the only residual connection. Dropout (0.1, as elsewhere in the model) is applied to $h_t$ entering $\delta_\psi$, but not to the skip path.
Alongside the usual next-token cross-entropy $\mathcal{L}_{\mathrm{NTP}}$, two auxiliary losses compare that prediction against the true $h_{t+1}$: a Smooth~L1 loss $\mathcal{L}_{h}$ on the hidden state itself, and a token-space divergence $\mathcal{L}_{\mathrm{KL}} = D_{\mathrm{KL}}\big(p_\theta^{\mathrm{sg}}(\cdot \mid \mathrm{sg}[h_{t+1}]) \,\|\, p_\theta^{\mathrm{sg}}(\cdot \mid \hat{h}_{t+1})\big)$. Targets are detached ($\mathrm{sg}[\cdot]$) to prevent representational collapse, and the LM head is frozen inside the KL term, so the only remaining gradient path runs through the predicted state $\hat{h}_{t+1}$, shaping the backbone's representations rather than the LM head. The total objective is
\begin{equation}
\mathcal{L} = \mathcal{L}_{\mathrm{NTP}} + \lambda_{h}\,\mathcal{L}_{h} + \lambda_{\mathrm{KL}}\,\mathcal{L}_{\mathrm{KL}},
\label{eq:nextlat}
\end{equation}
with $\lambda_{h} = 1.0$ and $\lambda_{\mathrm{KL}} = 0.5$ throughout; an auxiliary cross-entropy on the predicted state's emissions is available but carries weight zero in every reported run. We use the one-step setting, which is what the belief-state result requires; the original NextLat paper additionally studies multi-step rollouts in service of speculative decoding. Unlike the token-level losses, $\mathcal{L}_{h}$ applies at every position whose prediction does not cross a document boundary, so belief-state structure is shaped even while processing context. These losses encourage the hidden states to converge toward a \emph{belief state}, a sufficient statistic of the past for predicting the future. The auxiliary dynamics model is discarded after training, preserving the unmodified base architecture and standard autoregressive inference.

\section{Pretraining and Evaluation}
\label{method}

\subsection{Challenge Setting and Corpora}
\label{data}

We train on the BabyLM strict-small and strict tracks, which provide 10 million words and 100 million words respectively. Both tracks consist of developmentally plausible English text drawn from six sources: CHILDES, OpenSubtitles, Simple Wikipedia, Gutenberg, BNC spoken, and Switchboard \citep{choshen2026babylmturns4papers,warstadt-etal-2023-findings}. The strict-small regime is the most data-limited BabyLM setting and is therefore the most diagnostic of inductive-bias effects.

We tokenize with byte-pair-encoding (BPE) tokenizers in the GPT-BERT recipe \citep{charpentier-samuel-2024-bert}, each with a vocabulary of 16{,}384. The strict-small track and the 18-layer strict model use the tokenizers distributed with the official BabyLM GPT-BERT baselines (the 10M-word and 100M-word releases, respectively); the 16-layer wide (1024 hidden dim) strict model uses a 2026 retraining of the 100M-word tokenizer on the 2026 strict corpus. Training sequences are packed to a length of 512 tokens (264 for the symbol-retrieval comparison; Appendix~\ref{app:configs}). To preserve the short-utterance structure of child-directed input, we segment sequences at end-of-sequence (EOS) boundaries: tokens after an EOS boundary cannot attend to tokens before it, and the auxiliary dynamics model is never penalized for predicting latent states across EOS: causal next-token prediction (NTP) training with utterance-level segmentation.

\subsection{Model Configurations and Pretraining}
\label{config}

All strict-small models share a 12-layer, 768-dimensional decoder-only backbone with a BPE vocabulary of 16{,}384, rotary positional encodings, pre-LayerNorm blocks, and GELU or SwiGLU feed-forward activations depending on configuration. The standard Transformer baseline uses 12 self-attention heads; DAT replaces these with $n_h^{\mathrm{sa}}$ sensory and $n_h^{\mathrm{ra}}$ relational heads, with the split controlling the sensory/relational allocation. LM-head tying varies independently of the objective: NextLat runs use an untied head, matching the source NextLat setup, while NTP runs appear with both tied and untied heads. Tied models have $\approx$123.3M parameters and untied models $\approx$135.9M, the untied head adding $\approx$12.6M parameters (the auxiliary NextLat dynamics model adds $\approx$5.9M parameters and is discarded after training). All models train for 10 epochs with gradient clipping at 1.0. Batch sizes, learning rates, and head splits vary across the three analysis families and are summarized in Appendix~\ref{app:configs}. The strict-track submissions use the Muon/LambW recipe of Section~\ref{optimization}, and their full specifications are in Appendix~\ref{app:specs}.

\subsection{Optimization}
\label{optimization}

Our experiments use two optimization recipes. The strict-small analysis families use AdamW \citep{loshchilov2019adamw} with no weight decay, $\beta_1{=}0.9$, $\beta_2{=}0.999$, and the family-specific learning rates of Appendix~\ref{app:configs}. The strict-track submissions use Muon \citep{jordan2024muon,liu2025muonscalable} for matrix-valued hidden-layer parameters and LambW (our own decoupled-weight-decay variant of LAMB \citep{you2020lamb}, which relates to LAMB as AdamW relates to Adam) for embeddings, the LM head, symbol tables, normalization parameters, and biases. Muon applies a Newton--Schulz approximation to orthogonalize momentum updates; we use a Muon learning rate of 0.02, momentum of 0.95, and 5 Newton--Schulz steps. LambW retains LAMB's layerwise trust ratio while applying weight decay independently of the adaptive update, following the decoupling principle of AdamW \citep{kingma2014adam,loshchilov2019adamw}. We use a LambW learning rate of $10^{-3}$ and weight decay of 0.01. We compare the two recipes head-to-head in one controlled strict-track experiment, with AdamW at a learning rate of $5\times10^{-4}$ (Table~\ref{tab:recipe_comparison}, Appendix~\ref{app:results}); because it jointly changes optimizer, learning rate, and weight decay, the comparison reflects the whole recipe rather than the optimizer alone.

\subsection{Experimental Comparisons}
\label{experiments-design}

Our experimental matrix comprises 140 training/evaluation rows across 47 experiment groups, organized into three analysis families: (i)~an architecture comparison (standard Transformer vs.\ DAT) with 77 rows, (ii)~a complete RCA/SwiGLU symbol-retrieval comparison at 264-token context window with 35 rows (7 symbol conditions $\times$ 5 seeds), and (iii)~an RCA SA/RA head-ratio sweep with 28 rows (7 ratios at 512-token context with 3 seeds, plus one 264-token run per ratio). All models use the strict-small (10M-word) track. SwiGLU was chosen early in an unreported comparison with GELU, and was held fixed for all the submitted models, as well as the symbol-retrieval and head-ratio comparison families; the architecture comparison includes both GELU and SwiGLU runs, with activation partly coupled with symbol mechanism.

The architecture comparison varies four factors in a partially crossed design: \textbf{Base architecture}---standard self-attention with 12 heads vs.\ DAT with 9 sensory + 3 relational, or 6 + 6 heads; \textbf{Training objective}---next-token prediction (NTP) vs.\ Next-Latent Prediction (NextLat); \textbf{LM head}---tied vs.\ untied; and \textbf{Relational attention type}---full relational attention (RA), relational cross-attention (RCA), or disentangled RCA (DisRCA).

\subsection{Evaluation}
\label{stats}

We evaluate on the BabyLM zero-shot benchmark suite, which probes distinct linguistic and cognitive dimensions without task-specific fine-tuning: grammatical phenomena (\textbf{BLiMP}, \textbf{BLiMP supplement}, \citealt{blimp}), conceptual property knowledge (\textbf{COMPS}, \citealt{misra-etal-2023-comps}), discourse tracking and world knowledge (\textbf{Entity Tracking}, \citealt{kim-schuster-2023-entity}, \textbf{EWoK}, \citealt{ivanova2025elementsworldknowledgeewok}, \textbf{GlobalPIQA}, \citealt{chang2025globalpiqa}), and human-likeness measures (fit to human reading-time and EEG data, and age of acquisition). We also evaluate selected models on the \textbf{(Super)GLUE} fine-tuning suite. Detailed benchmark definitions are provided in Appendix~\ref{app:benchmarks}.

Our claims rest on two statistical models, each fit on the strict-small (10M-word) track.

\paragraph{Binomial GLMMs on BLiMP.}
We fit binomial generalized linear mixed-effects models (GLMMs) on BLiMP pseudo-items (successes out of items per pseudo-item). A baseline-vs-DAT model compares the standard Transformer against the 9SA/3RA RCA/SwiGLU DAT configuration, with fixed effects for base architecture, LM-head tying, and training objective; an internal-DAT model adds fixed effects for relational attention type, head split, feed-forward activation and symbol-position setting, and context length. All models use Type~III Wald $\chi^2$ tests with sum-to-zero contrast coding, and crossed random intercepts for experiment (which identifies the seed and configuration), BLiMP subtest, and pseudo-item~ID. Post-hoc contrasts use Tukey-adjusted pairwise comparisons. Throughout, the architecture comparison's factors are only partially crossed (tied/NextLat is unobserved), so contrasts involving a missing cell are model-based extrapolations.

\paragraph{Reading regression.}
A cross-experiment linear mixed-effects model regresses the incremental $R^2$ ($\Delta R^2$) gained by adding model surprisal to baseline reading-time predictors, with crossed random intercepts for experiment and reading measure, and fixed effects for training objective, model type, LM-head tying, context length, and reading condition. The last of these is an evaluation-condition control: each experiment contributes two $\Delta R^2$ rows per measure, one from the base reading regression and one from a spillover regression that adds previous-word predictors.

\section{Experiments}
\label{results}

\subsection{BabyLM Submission}
\label{res:strict}

Table~\ref{tab:strict_results} summarizes our strict-track models and the official GPT-2 baseline under the official leaderboard evaluation. Our two strongest entries are DAT models trained with NextLat, the Muon/LambW recipe, and RoPE-based relative symbols, each on a two-phase curriculum that starts with line (utterance) EOS segmentation and switches to document EOS boundaries: an 18-layer model (768-dim, 9SA/3RA, RCA) trained with a curriculum of 6 line-EOS / 4 document-EOS epochs, and a 16-layer wide model (1,024-dim, 12SA/4RA, RA) trained with a curriculum of 2 line-EOS / 8 document-EOS epochs. Both submitted checkpoints are exponential moving averages (EMA) of the training weights; Appendix~\ref{app:specs} gives their full specifications. At the time of writing our two strongest entries rank 6th and 7th of 55 overall; our best, the 16-layer wide (1024-dimensional) model, also ranks 3rd of 55 on the NLP-task subset, which comprises all tasks except age of acquisition and fit to reading data. The wide model beats the baseline on eight of the nine benchmarks (all but GlobalPIQA) and holds the highest EWoK score on the strict-track leaderboard at the time of writing (59.54); the 18-layer model beats the baseline on seven of nine, and shows no reliable age-of-acquisition correlation (0.00; non-significant correlations are scored as zero, Appendix~\ref{app:benchmarks}) whereas the baseline's correlation is significantly negative ($-11.58$). Its curriculum trades a little BLiMP accuracy for a large gain on the BLiMP supplement ($+8.5$ points over the non-curriculum 18L model). Note that these are not controlled parameter or training-recipe matched comparisons: the GPT-2 baseline has 98M parameters trained with AdamW, while our entries range from 123M to 304M parameters trained with Muon/LambW. A parameter-matched architecture comparison is provided in Section~\ref{res:arch} below with models trained on the strict-small track.

On the strict-small (10M-word) track, we submit two 12-layer DAT models with the same 9SA/3RA RCA architecture, SwiGLU activations, RoPE-based relative symbols, and Muon/LambW recipe: a base NTP model with a tied head (123M parameters) and a NextLat model with an untied head (136M parameters), both trained for 10 epochs at 512-token context with seed 1. At the time of writing they rank 60th and 53rd of 127 entries by Overall Average (37.91 and 38.16, versus 37.38 for the GPT-2 baseline), but place near the top on the structural and knowledge-oriented benchmarks: the NextLat model ranks 8th of 127 on BLiMP (72.34, versus 65.23 for the baseline) and 6th on EWoK (52.93). Relative to the base model, NextLat improves BLiMP ($+1.9$), EWoK ($+2.7$), reading (7.12 vs.\ 6.56), and (Super)GLUE (63.97 vs.\ 62.96); the two models differ in LM-head tying and therefore parameter count. Table~\ref{tab:strict_small_results} in Appendix~\ref{app:results} reports the full strict-small results.

\begin{table*}[t]
\centering\footnotesize
\setlength{\tabcolsep}{3.5pt}
\begin{tabular}{@{}lrrrrrrrrrrr@{}}
\toprule
Model & BLiMP & Supp. & EWoK & Ent.\ trk. & COMPS & PIQA & GLUE$^{*}$ & Reading & AoA & Overall & NLP \\
\midrule
\shortstack[l]{GPT-2 baseline\\(12L)} & 74.73 & 65.00 & 54.37 & 16.91 & 55.85 & 36.62 & 67.75 & 6.93 & $-11.58$ & 40.73 & 53.03 \\
\addlinespace
\shortstack[l]{DAT 12L\\(NTP)} & 79.81 & 60.03 & 56.66 & 20.84 & 57.84 & \textbf{37.65} & 65.92 & 5.76 & $-12.02$ & 41.39 & 54.11 \\
\shortstack[l]{DAT 18L\\(NextLat)} & \textbf{80.62} & 62.01 & 56.99 & \textbf{20.93} & 58.36 & 34.72 & 68.90 & 6.49 & 0.00 & 43.23 & 54.65 \\
\shortstack[l]{DAT 18L\\(NextLat, curric.)} & 78.94 & 70.51 & 57.25 & 20.47 & 58.69 & 32.72 & 69.95 & 6.22 & 0.00 & 43.86 & 55.50 \\
\shortstack[l]{DAT 16L wide\\(NextLat, curric.)} & 79.49 & \textbf{70.96} & \textbf{59.54} & 20.89 & \textbf{59.22} & 36.17 & \textbf{71.43} & \textbf{7.10} & $-9.48$ & \textbf{43.92} & \textbf{56.81} \\
\bottomrule
\end{tabular}
\caption{Strict-track (100M-word) results from the official BabyLM leaderboard at the time of writing (55 entries). All DAT models use relative symbol retrieval with RoPE-based relative symbols (no learned symbol library; Section~\ref{dat:symbols}), SwiGLU activations, and the Muon/LambW recipe: the 12L model has 123M parameters (9SA/3RA, RCA, 768-dim, tied head, seed 0), the 18L models have 191M parameters (9SA/3RA, RCA, 768-dim, seed 1), and the 16L wide model has 304M parameters (12SA/4RA, RA with four relation channels, 1,024-dim, seed 1). Full specifications of the two top entries are in Table~\ref{tab:strict_model_specs}. PIQA is GlobalPIQA; GLUE$^{*}$ is the leaderboard's (Super)GLUE fine-tuning average; Reading is the leaderboard's composite human-likeness score, which aggregates eleven eye-tracking, self-paced-reading, and ERP measures (Appendix~\ref{app:benchmarks}); AoA is the age-of-acquisition score (Appendix~\ref{app:benchmarks}); Overall is the leaderboard Overall Average (equal weight across the nine benchmarks); NLP is the Overall Average restricted to the seven NLP-task benchmarks, excluding the two human-likeness measures (reading and age of acquisition, which form the leaderboard's Human-like Average). The GPT-2 baseline is the official challenge baseline (98M parameters, AdamW). Bold marks the best value per column. The comparison with the GPT-2 baseline is not parameter or training-recipe matched.}
\label{tab:strict_results}
\end{table*}

\subsection{Does Dual Attention Help?}
\label{res:arch}

For structural syntactic reasoning (BLiMP), architecture choice is the dominant factor. A binomial GLMM with fixed effects for base architecture (standard Transformer vs.\ DAT), LM-head tying, and training objective reveals a large and significant main effect of architecture ($\chi^2(1)=139.43$, $p<0.001$; Table~\ref{tab:anova_baseline_vs_dat}, Appendix~\ref{app:results}). Estimated contrasts show that DAT scores higher than the standard Transformer baseline across all four head-tying $\times$ objective conditions, with advantages ranging from $+0.083$ to $+0.288$ on the log-odds scale (all $p<0.001$; Table~\ref{tab:contrasts_baseline_vs_dat}). Note that contrasts at factor combinations not present in the training matrix (such as tied/NextLat) are model-based extrapolations (Section~\ref{stats}). The best DAT configuration reaches $70.66\%$ BLiMP against $68.16\%$ for the best standard Transformer (Table~\ref{tab:main_results}, Appendix~\ref{app:results}).

\begin{figure}[t]
  \centering
  \includegraphics[width=0.95\columnwidth]{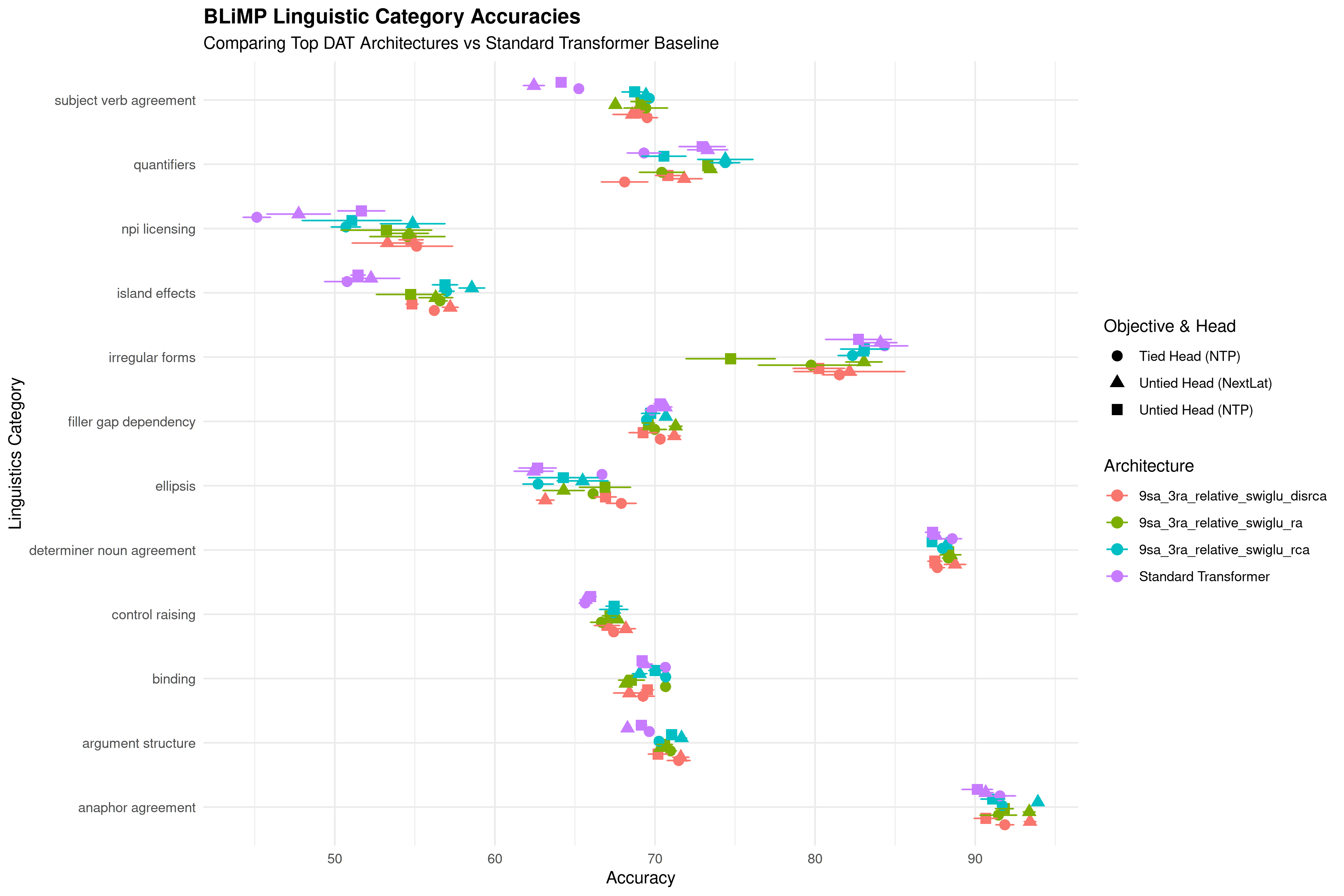}
  \caption{BLiMP accuracy by linguistic category, comparing DAT and standard Transformer configurations. Gains are largest on structural domains (e.g., island effects, subject--verb agreement).}
  \label{fig:blimp_ling_terms}
\end{figure}

The advantage is not uniform across linguistic phenomena. A focused architecture $\times$ linguistic-domain interaction model shows a significant interaction ($\chi^2(12)=573.07$, $p<0.001$): DAT yields larger gains on structural domains such as island effects, subject--verb agreement, and quantifiers than on lexical or semantic ones. Figure~\ref{fig:blimp_ling_terms} shows BLiMP accuracy by linguistic category for DAT and standard Transformer configurations; Figure~\ref{fig:blimp_means} in Appendix~\ref{app:results} gives mean BLiMP accuracy across configurations.

\begin{table}[t]
\centering\small

\resizebox{\columnwidth}{!}{%
\begin{tabular}{@{}lllrrrr@{}}
\toprule
Contrast & LM head & Objective & Estimate & SE & $z$ & $p$ \\
\midrule
baseline $-$ DAT & tied   & NextLat & $-0.288$ & 0.038 & $-7.49$ & $<0.001$ \\
baseline $-$ DAT & untied & NextLat & $-0.186$ & 0.022 & $-8.37$ & $<0.001$ \\
baseline $-$ DAT & tied   & NTP     & $-0.185$ & 0.022 & $-8.33$ & $<0.001$ \\
baseline $-$ DAT & untied & NTP     & $-0.083$ & 0.022 & $-3.72$ & $<0.001$ \\
\bottomrule
\end{tabular}%
}
\caption{Estimated contrasts (baseline $-$ DAT) from the BLiMP GLMM, by LM-head tying and training objective. Negative estimates indicate that DAT scores higher than the standard Transformer baseline. All four contrasts are significant at $p<0.001$ (Tukey-adjusted).}
\label{tab:contrasts_baseline_vs_dat}
\end{table}

\subsection{Does NextLat Complement Dual Attention?}
\label{res:nextlat}

While architecture governs structural accuracy, the training objective governs alignment with human cognitive processing. A cross-experiment reading regression models the incremental $R^2$ ($\Delta R^2$) gained by adding model surprisal to baseline reading-time predictors. The training objective is significant ($t=5.22$, $p<0.001$): NextLat models explain more variance in human reading measures than NTP models (Table~\ref{tab:reading_regression}, Appendix~\ref{app:results}). Model type (DAT vs.\ standard) is also significant ($t=-2.61$, $p=0.010$), with standard Transformers explaining slightly more reading-time variance than DAT models, and context length is significant ($t=-7.08$, $p<0.001$).
 The matrix separates the objective from LM-head tying: its untied/NTP and untied/NextLat cells are matched on both, and NextLat gains 0.83 BLiMP points across them (Appendix~\ref{app:results}).

\paragraph{Fine-tuned task performance.}
On the strict-track (Super)GLUE benchmarks, NextLat improves 5 of 7 task accuracies, with the largest gains on MultiRC ($+5.7$ percentage points) and WSC ($+3.8$; Table~\ref{tab:finetune_results}, Appendix~\ref{app:results}). These two strict-track models differ in LM-head tying and parameter count ($\approx 12.6$M).

\paragraph{Architecture$\times$objective interaction.}
The full baseline-vs-DAT model reveals significant interactions between base architecture and both head tying ($\chi^2(1)=10.62$, $p=0.001$) and training objective ($\chi^2(1)=10.81$, $p=0.001$): the size of the DAT advantage depends on the objective and on whether the LM head is tied, so the two interventions are not simply additive.

\subsection{Which Relational Mechanisms Matter?}
\label{res:rca}

The RCA/SwiGLU 264-token symbol-retrieval comparison is the only fully-powered complete subset in our matrix: seven symbol conditions $\times$ five seeds (Table~\ref{tab:rca_symbols}, Appendix~\ref{app:results}). A binomial GLMM on BLiMP subtests shows a large, significant effect of the symbol-retrieval mechanism ($\chi^2(6)=134.29$, $p<0.001$).

Contrasts against the learned-relative-symbol baseline (Table~\ref{tab:contrasts_rca_symbols}) show that the two relational-symbolic conditions are significantly worse on BLiMP than the learned-relative baseline's $69.34\%$: \texttt{relsymbolic\_n4}, which combines the larger size-4 relational-symbol neighborhood with a learned symbol library, at $66.32\%$ ($p<0.001$), and \texttt{relsymbolic} at $68.33\%$ ($p=0.023$). The remaining five conditions (learned-relative, RoPE-relative, learned-positional, sinusoidal-positional, and symbolic-attention) are not significantly different from one another (all pairwise $p > 0.5$, Tukey-adjusted); beyond the relational-symbolic disadvantage, the comparison does not support a finer ranking among symbol sources. That indifference is itself useful: RoPE-based relative symbols match the learned library they replace ($69.43\%$ vs.\ $69.34\%$) while eliminating its parameter cost---a learned relative-symbol library holds $(2\Delta{+}1)\,d$ parameters, 1.05M for the submitted 1{,}024-dimensional models with $\Delta{=}512$---and its maximum-offset table.

\subsection{Which Relational Attention Type?}
\label{res:ratype}

The controlled RA-type comparison varies only attention type, objective, and LM-head tying across 25 runs, holding architecture, symbols, context, and recipe fixed (Appendix~\ref{app:results}). On BLiMP, the three types are indistinguishable: the internal-DAT GLMM finds no significant effect of attention type ($\chi^2(2)=3.50$, $p=0.174$), and the largest between-type spread in any condition is 0.76 points against a pooled within-cell seed standard deviation of 0.54. RCA is numerically highest in all three conditions. This converges with the DAT paper's relational-games ablation, where RA and RCA perform similarly \citep{altabaa2025disentangling}, and contradicts its language-modeling ablation, in which RCA-head DAT performs ``no better than a standard Transformer with a matching total number of heads'': ours beats exactly that baseline by 2.5 BLiMP points (Section~\ref{res:arch}). At 10M words the simpler RCA, which routes only symbols and needs no relation-projection parameters, is the equal of full RA.

At 100M words the ordering changes. In a matched 18-layer pair differing only in attention type, RA beats RCA on six of seven benchmarks (BLiMP 79.30 vs.\ 78.44, EWoK $+1.7$, entity tracking $+3.7$). RA is also the least stable: in a controlled 12-layer, 1,024-dimensional pair its training loss diverged and never recovered below its epoch-0 value, while RCA trained monotonically to 79.72 BLiMP and DisRCA to 79.38 (Appendix~\ref{app:results}). The extra capacity RA carries---separate relation projections and four relation channels---appears to need both data and a stable recipe.

\subsection{SA/RA Head-Ratio Sweep}
\label{res:sara}

We sweep the sensory-to-relational head ratio at 512-token context, with 3 seeds per ratio (Table~\ref{tab:sa_ra_ratio}, Appendix~\ref{app:results}). A mixed-effects model with seed as a random intercept shows that higher relational-attention (RA) fraction significantly decreases BLiMP accuracy (slope = $-1.827$, $t = -5.887$, $p < 0.001$) and EWoK accuracy (slope = $-1.039$, $p = 0.021$). With 12 heads in total the two fractions are complements: a larger relational stream is also a smaller sensory one. The RA fraction has no significant effect on COMPS ($p = 0.201$) or reading measures ($p = 0.488$), and COMPS and supplement scores stay flat across ratios, so the head-ratio allocation acts mainly on syntactic structures. Mean BLiMP peaks at the balanced 6SA/6RA split ($70.33\%$), but 9SA/3RA --- the allocation used in all submitted models --- trades 0.6 BLiMP points for a 1.4-point supplement gain and the best COMPS score.

\section{Related Work}
\label{relwork}

\paragraph{BabyLM and data-efficient language models.}
The BabyLM challenge \citep{warstadt-etal-2023-findings,choshen2026babylmturns4papers} provides a shared evaluation for sample-efficient pretraining on developmentally plausible corpora. Successful entries have mostly intervened on the data or the training objective rather than on the attention mechanism: curricula and corpus filtering, distillation, and hybrid objectives such as GPT-BERT \citep{charpentier-samuel-2024-bert}, which merges causal and masked prediction in one Transformer stack. Architecture-level interventions like our submission evaluated under the full suite are comparatively rare.

\paragraph{Relational bottleneck and dot-product relations.}
The closest point of comparison to our work is the Abstractor architecture and relational cross-attention (RCA) \citep{altabaa2024abstractors} and DAT \citep{altabaa2025disentangling}. While DAT greatly improves data efficiency on purely relational tasks, the evidence for its usefulness in language modeling is a modest perplexity gain on web text (16.94 to 16.09 at the 350M scale, and less at larger scales), with no evaluation of what the relational stream does to linguistic generalization. Our work extends the original DAT design with new symbol-retrieval mechanisms and studies them in a developmentally plausible training regime with controlled comparisons under the BabyLM evaluation suite.

The motivation for RCA and DAT connects to older work on compositionality and variable binding \citep{smolensky1990tensor,holyoak2000proper}. The ESBN \citep{esbn} shows that a recurrent network augmented with an external memory can implement variable binding and indirection, allowing symbol-like representations to emerge and enabling near-perfect generalization of abstract rules to novel entities; CoRelNet \citep{kerg2022neural} simplifies this mechanism, showing that a matrix of dot-product similarities can itself serve as a relational representation that supports out-of-distribution generalization. A broader cognitive framework for all of these works is provided in the relational bottleneck proposal \citep{webb2024relational}, which argues that architectures can benefit from restricting part of the computation to relations rather than object-specific sensory content.

\paragraph{Latent-state objectives and cognitive/psycholinguistic evaluation.}
Next-Latent Prediction \citep{teoh2026nextlatent} was introduced and evaluated as a world-modeling objective, on world-model benchmarks and on the latent rollouts that support speculative decoding. Whether the compression pressure it imposes also makes a model more human-like is a separate question, and the one our reading regression asks (Section~\ref{res:nextlat}). It is the same pressure the Now-or-Never bottleneck attributes to human comprehension \citep{christiansen2016now}, and the link between a model's word-by-word predictions and human reading effort is the standard currency for testing it \citep{hale2001probabilistic,lewis2005activation}.

\section{Conclusion}
\label{conclusion}

We set out to ask whether the data efficiency shown by relational attention on purely relational tasks transfers to language modeling. Our answer is a qualified yes. Under the strict-small regime, architecture is the dominant factor for structural linguistic generalization, worth roughly 2.5 BLiMP points between the best configurations of each at matched depth, width, and head count; the gain comes from adding a relational stream rather than from enlarging it, and at this scale the three relational mechanisms are interchangeable. The NextLat objective is secondary but significant, and interacts with architecture: NextLat models track human reading measures more closely and improve 5 of 7 fine-tuned (Super)GLUE accuracies. Among symbol sources only the relational-symbolic variants underperform, and our novel parameter-free RoPE-based relative symbols match the learned library they replace. At 100M words full relational attention pulls ahead, and our two strongest models rank among the top strict-track entries at the time of writing.
\label{endofbody}

\pagebreak

\section*{Limitations}

\paragraph{Statistical caveats.}
The binomial GLMMs model aggregated per-item counts without observation-level random effects, so overdispersion is not addressed and exact $p$-values may be anti-conservative.

\paragraph{Incomplete replication.}
Most experiment groups contain only 3 seeds; only the RCA/SwiGLU 264-token symbol-retrieval comparison has the full 5-seed design.

\paragraph{NextLat effect size.}
While the training objective is statistically significant for cognitive alignment ($t=5.22$, $p<0.001$ in the reading regression), the effect is secondary to the architecture effect on structural generalization. The architecture$\times$objective interaction ($\chi^2(1)=10.81$, $p=0.001$) indicates that the NextLat benefit is not uniform across architectures. The evidence for the NextLat inductive bias driving structural generalization, while statistically significant, does not have the same weight as architecture in our experiments.

\paragraph{Scope.}
All replicated results use the strict-small (10M-word) track. Our submitted strict-track entries are single-seed; the recipe-comparison models have two (NextLat) and three (NTP) seeds. At strict scale the base and NextLat runs also differ in LM-head tying and parameter count; the strict-small matrix separates the two, since its untied/NTP and untied/NextLat cells are matched on both. Seed spread is non-trivial: the Muon/LambW NTP trio spans 79.81/79.20/78.78 BLiMP (range 1.03) and 56.66/55.23/55.35 EWoK (range 1.43); the NextLat pair spans 79.75/79.03 BLiMP (range 0.72). The seed ranges exceed the NTP-vs-NextLat gaps the recipe table reports (0.06 BLiMP, 0.51 EWoK on seed 0). Entity tracking has one extreme outlier: a third NTP seed scores 33.63 against 20.84 and 20.80 for the other two. The comparison with the official GPT-2 baseline is likewise not parameter- or recipe-matched. We do not evaluate on the strict (100M-word) track with full replication, leaving open whether the DAT architecture and NextLat objective effects scale with more data. The standard Transformer baseline uses 12 self-attention heads; we do not report a parameter-matched baseline that scales head count independently of the sensory/relational split. The architecture comparison is also not matched on every other axis: the DAT and standard configurations differ in feed-forward activation, initialization, and symbol mechanism as well as in attention.

\paragraph{Dataset and task coverage.}
We evaluate on the BabyLM strict-small zero-shot suite (BLiMP, BLiMP supplement, COMPS, Entity Tracking, EWoK, and the human-likeness measures) and the strict-track (Super)GLUE fine-tuning suite. We do not report generation quality. The human-likeness measures derive from a single resource of 205 sentences \citep{devarda2023cloze} covering eye-tracking, self-paced reading, and EEG responses.

\paragraph{Implementation faithfulness.}
The DAT implementation in our experiments differs from the original DAT formulation in several respects: the use of SwiGLU vs.\ GELU feed-forward activations, tied vs.\ untied LM heads, the symbol-retrieval mechanism (RoPE-based relative symbols rather than a learned symbol library), and the weight initialization scheme. The relation activation used in RCA (identity or sigmoid rather than softmax) is a hyperparameter setting the original design provides for (in principle), but has remained previously underexplored.

\paragraph{Optimizer-recipe comparison.}
The Muon/LambW recipe improves all five accuracy-based zero-shot measures (BLiMP $+1.73$pp, BLiMP supplement $+1.15$pp, EWoK $+1.77$pp, entity tracking $+1.99$pp, COMPS $+1.02$pp; Table~\ref{tab:recipe_comparison}), while reading-time measures decrease slightly. This comparison jointly changes optimizer family, learning rate, and weight decay, so the improvement is a property of the recipe and (potentially) not of specific components in isolation.

\section*{Acknowledgments}
AB gratefully acknowledges the compute support from the National Research Platform (NRP) at the University of California, San Diego. NRP has been developed, and is supported in part, by funding from National Science Foundation, from awards 1730158, 1540112, 1541349, 1826967, 2112167, 2100237, and 2120019, as well as additional funding from community partners \citep{weitzel2025nrp}. ET and JD were supported by the European Research Council (ERC), grant 101088098 - MEMLANG. Views and opinions expressed are however those of the authors only and do not necessarily reflect those of any funding agency. Part of this work also used resources available through the Dutch national e-infrastructure with the support of the SURF Cooperative using grant no. EINF-18356.

LLM assistants were used in circumscribed, closely supervised ways during code development and the editing process of the manuscript. On the spectrum between local, narrow autocomplete vs.\ global, large-scale LLM-based generation of code or text, this work stays decidedly close to the local / narrow end of the spectrum. We retain full responsibility for all content, and all errors are our own.

\bibliography{custom}

\appendix

\section{Further Details}
\label{app_dets}

Our code---model implementation, training and evaluation scripts, configurations, and analysis code---is available at \url{https://github.com/abrsvn/babylm_dat_2026}. Our submitted checkpoints are also publicly available on the HuggingFace Hub and are linked from the official leaderboard entries.

\section{Strict-Small Training Configurations}
\label{app:configs}

Table~\ref{tab:strict_small_configs} summarizes the training configurations of the three strict-small analysis families (Section~\ref{experiments-design}), referenced from Section~\ref{config}.

\begin{table}[t]
\centering\footnotesize
\setlength{\tabcolsep}{2.5pt}
\begin{tabular}{@{}lccc@{}}
\toprule
 & Architecture & Symbols & Head ratio \\
\midrule
Context & 512 & 264 & 512 \\
Batch size & 12 or 16 & 16 & 12 \\
LR (AdamW) & $5\times10^{-5}$ & $10^{-3}$ & $5\times10^{-5}$ \\
SA/RA split & 6/6, 9/3 & 9/3 & 1/11 to 11/1 \\
Attn type & RA, RCA, DisRCA & RCA & RCA \\
Rel.\ activation & identity & sigmoid & identity \\
Activation & GELU/SwiGLU & SwiGLU & SwiGLU \\
LM head & tied/untied & tied & tied \\
Seeds & 3 & 5 & 3 \\
\bottomrule
\end{tabular}
\caption{Training configurations of the three strict-small analysis families (Section~\ref{experiments-design}). SA/RA = sensory/relational head split. All runs use AdamW, 10 epochs, gradient clipping at 1.0, and 12-layer, 768-dimensional backbones. The architecture comparison also spans training objective (NTP vs.\ NextLat) and includes single auxiliary runs at 264- and 1,280-token context (the context-length factor in the internal-DAT analysis of Appendix~\ref{app:results}); LM-head tying varies independently of the objective: NextLat runs are untied, NTP runs appear both ways.}
\label{tab:strict_small_configs}
\end{table}

\section{Top Strict-Track Model Specifications}
\label{app:specs}

Table~\ref{tab:strict_model_specs} gives the full model and training specifications of our two strongest strict-track entries (Section~\ref{res:strict}). Parameter counts are computed directly from the published checkpoints; the remaining settings are taken from the training configurations. Both models use the RoPE-based relative symbols introduced in Section~\ref{dat:symbols}, which add no learned parameters, and both submitted checkpoints are exponential moving averages of the training weights.

\begin{table*}[t]
\centering\footnotesize
\begin{tabular}{@{}lp{5.6cm}p{5.6cm}@{}}
\toprule
Setting & 18L curriculum model & 16L wide model \\
\midrule
Leaderboard name & DAT Strict Curriculum & DAT Strict NextLat Final \\
Parameters & 191.2M & 304.3M \\
Layers / hidden dim. & 18 / 768 & 16 / 1,024 \\
Attention heads (SA/RA) & 9 / 3 & 12 / 4 \\
Attention type & RCA & RA (4 relation channels, identity activation) \\
Symbol mechanism & \multicolumn{2}{p{11.4cm}}{RoPE-based relative symbols, shared across layers; no learned symbol library (parameter-free)} \\
Feed-forward & SwiGLU, hidden dim.\ 3,072 & SwiGLU, hidden dim.\ 4,096 \\
Normalization & \multicolumn{2}{p{11.4cm}}{pre-LayerNorm} \\
Positional encoding & \multicolumn{2}{p{11.4cm}}{RoPE ($\theta{=}10{,}000$); max.\ relative position 512} \\
LM head & \multicolumn{2}{p{11.4cm}}{untied} \\
Vocabulary / sequence length & \multicolumn{2}{p{11.4cm}}{16,384 / 512 packed tokens (514 model positions)} \\
Tokenizer & GPT-BERT 100M (2024 release) & GPT-BERT 100M (2026 release) \\
Batch & 48 sequences of 512 tokens & 36 sequences of 512 tokens \\
Random seed & 1 & 1 \\
Curriculum & Phase A: 6 epochs line-EOS; Phase B: 4 epochs document-boundary & Phase A: 2 epochs line-EOS; Phase B: 8 epochs document-boundary \\
Muon lr (Phase A / Phase B) & 0.02 / 0.008 & 0.02 / 0.005 \\
LambW lr (Phase A / Phase B) & $10^{-3}$ / $4\times10^{-4}$ & $10^{-3}$ / $2.5\times10^{-4}$ \\
Weight decay & 0.01 & 0.05 \\
LR schedule & \multicolumn{2}{p{11.4cm}}{1\% warmup, cosine decay to 10\% of the initial rate} \\
NextLat & \multicolumn{2}{p{11.4cm}}{horizon 1; MSE weight 1.0, KL weight 0.5, CE weight 0.0} \\
EMA & \multicolumn{2}{p{11.4cm}}{decay 0.999; submitted weights are the EMA checkpoints} \\
Initialization / dropout & \multicolumn{2}{p{11.4cm}}{scaled-normal (0.02) projections / 0.1} \\
\bottomrule
\end{tabular}
\caption{Full specifications of our two strongest strict-track models, verified against the published checkpoints and training configurations. Both are decoder-only DAT models trained from scratch on the BabyLM strict (100M-word) corpus for 10 total epochs: Muon (momentum 0.95, 5 Newton--Schulz steps) optimizes hidden weight matrices and LambW ($\beta_1{=}0.9$, $\beta_2{=}0.999$, $\epsilon{=}10^{-8}$) optimizes embeddings, heads, symbol tables, norms, and biases. Phase B is warm-started from the Phase-A EMA weights. The two models use different releases of the GPT-BERT 100M-word tokenizer, both with a vocabulary of 16{,}384: the 2024 release is byte-identical to the tokenizer distributed with the official GPT-BERT baseline, and the 2026 release is a retraining of the same tokenizer recipe on the 2026 strict corpus (Section~\ref{data}). Training sequences are packed to 512 tokens as in Section~\ref{data}; the models allocate 514 positions: the 512 packed tokens, a leading BOS slot, and one appended final-target token that the one-step NextLat rollout conditions on (nextlat-free models allocate 513). Parameter counts are checkpoint (backbone) counts; the training-time NextLat dynamics module adds a further 10.5M parameters and is discarded after training. No teacher model, synthetic data, or multimodal input is used.}
\label{tab:strict_model_specs}
\end{table*}

Table~\ref{tab:strict_glue_results} reports the official per-task fine-tuned (Super)GLUE scores for the two top entries and the GPT-2 baseline.

\begin{table}[t]
\centering\small
\setlength{\tabcolsep}{4.5pt}
\begin{tabular}{@{}lrrr@{}}
\toprule
Task & GPT-2 base & 18L curric. & 16L wide \\
\midrule
BoolQ & 69.66 & 69.48 & \textbf{70.52} \\
MNLI & 60.76 & 63.00 & \textbf{66.26} \\
MRPC & 85.34 & 88.00 & \textbf{88.74} \\
MultiRC & 65.92 & 65.68 & \textbf{66.30} \\
QQP & 71.56 & 73.14 & \textbf{74.26} \\
RTE & 57.55 & 66.91 & \textbf{70.50} \\
WSC & \textbf{63.46} & \textbf{63.46} & \textbf{63.46} \\
\midrule
(Super)GLUE avg. & 67.75 & 69.95 & \textbf{71.43} \\
\bottomrule
\end{tabular}
\caption{Official fine-tuned (Super)GLUE task scores for the GPT-2 baseline and our two top strict-track entries (Table~\ref{tab:strict_results}), from the official leaderboard evaluation. Metrics follow the leaderboard's definitions: F1 for MRPC and QQP, accuracy for the rest---including MultiRC, which the BabyLM pipeline scores by accuracy rather than by SuperGLUE's F1a and exact match. Bold marks the best value per row.}
\label{tab:strict_glue_results}
\end{table}

\section{Benchmark Definitions}
\label{app:benchmarks}

\paragraph{BLiMP.}
The Benchmark of Linguistic Minimal Pairs \citep{blimp} consists of 67 minimal-pair paradigms testing grammatical phenomena such as subject--verb agreement, island effects, and filler--gap dependencies. We use aggregate BLiMP accuracy and a decomposition by linguistic category.

\paragraph{BLiMP supplement.}
The BLiMP supplement contains additional minimal-pair paradigms beyond the core 67, probing further syntactic and semantic phenomena.

\paragraph{COMPS.}
The Conceptual Minimal Pair Sentences benchmark \citep{misra-etal-2023-comps} tests property attribution and property inheritance. Each pair attributes a property to two concepts, one that has it and one that does not (\emph{A robin/\#penguin can fly}); the harder subsets restate the property of a novel concept introduced as a subordinate of the positive or negative concept (\emph{A wug is a robin/penguin. Therefore, a wug can fly}), which controls for memorization of the literal test phrase, and then add the negative concept back as a distractor.

\paragraph{Entity Tracking.}
This benchmark \citep{kim-schuster-2023-entity} evaluates whether a model can maintain and update the changing states and locations of entities across a discourse, testing a core cognitive prerequisite for narrative comprehension.

\paragraph{EWoK.}
The Elements of World Knowledge benchmark \citep{ivanova2025elementsworldknowledgeewok} evaluates basic conceptual world knowledge across domains such as spatial relations, physical states, and social interactions. Each example consists of minimal-pair sentences testing whether a model understands concepts that help model the world.

\paragraph{Reading and eye-tracking.}
The BabyLM human-likeness benchmark regresses a model's per-word surprisal against eleven psycholinguistic dependent variables drawn from \citet{devarda2023cloze}: four eye-tracking measures (first-fixation, first-pass, go-past, and right-bounded reading time), self-paced reading time, and six event-related potential components (ELAN, LAN, N400, P600, EPNP, PNP). The incremental $R^2$ ($\Delta R^2$) reported in Table~\ref{tab:reading_regression} is the additional variance in these measures explained by adding model surprisal to baseline predictors, averaged over the eleven variables.

\paragraph{GlobalPIQA.}
Global PIQA \citep{chang2025globalpiqa} is a physical commonsense-reasoning benchmark spanning over 100 language varieties, with a culture-specific non-parallel split and a shared, culture-agnostic parallel split. The Strict and Strict-Small tracks use the English subset, evaluated cloze-style: completions are chosen by length-normalized conditional log-probabilities; the leaderboard score is the mean of the parallel and non-parallel splits \citep{choshen2026babylmturns4papers}.

\paragraph{Age of acquisition (AoA).}
Following \citet{chang2022word}, the AoA evaluation tracks word surprisal across training checkpoints, derives a model age of acquisition for each word, and compares it with children's acquisition ages from CDI norms. The reported score is the Pearson correlation between model and child acquisition ages (scaled by 100); correlations that are not significant at $p \le 0.1$ are scored as zero.

\section{Further Results}
\label{app:results}

\paragraph{Strict-small leaderboard results.}
Table~\ref{tab:strict_small_results} reports the official strict-small (10M-word) leaderboard results for our two submissions and the GPT-2 baseline, complementing the strict-track results of Table~\ref{tab:strict_results} and discussed in Section~\ref{res:strict}.

\begin{table*}[t]
\centering\footnotesize
\setlength{\tabcolsep}{3.5pt}
\begin{tabular}{@{}lrrrrrrrrrr@{}}
\toprule
Model & BLiMP & Supp. & EWoK & Ent.\ trk. & COMPS & PIQA & GLUE$^{*}$ & Reading & AoA & Overall \\
\midrule
GPT-2 baseline (12L) & 65.23 & 57.25 & 50.63 & \textbf{19.10} & 51.81 & 35.09 & 63.80 & 5.63 & $\mathbf{-12.15}$ & 37.38 \\
\addlinespace
DAT 12L (NTP) & 70.40 & \textbf{57.94} & 50.25 & 18.62 & \textbf{52.79} & \textbf{35.64} & 62.96 & 6.56 & $-14.00$ & 37.91 \\
DAT 12L (NextLat) & \textbf{72.34} & 56.40 & \textbf{52.93} & 18.45 & 52.69 & 33.65 & \textbf{63.97} & \textbf{7.12} & $-14.15$ & \textbf{38.16} \\
\bottomrule
\end{tabular}
\caption{Strict-small (10M-word) results from the official BabyLM leaderboard at the time of writing (127 entries). Both DAT models use 9SA/3RA RCA, SwiGLU activations, RoPE-based relative symbols, and the Muon/LambW recipe (Muon lr 0.02, effective batch of 16 sequences of 512 tokens, seed 1, 10 epochs; the NextLat run reaches that batch as 8 sequences with 2 gradient-accumulation steps); the NTP model has a tied LM head (123M parameters) and the NextLat model an untied head (136M parameters). Column definitions as in Table~\ref{tab:strict_results}. Bold marks the best value per column. The comparison with the GPT-2 baseline (98M parameters, AdamW) is not parameter- or recipe-matched.}
\label{tab:strict_small_results}
\end{table*}

\paragraph{Optimizer-recipe comparison.}
Table~\ref{tab:recipe_comparison} reports the strict-track optimizer-recipe comparison between AdamW and Muon/LambW at 12 layers, referenced in Sections~\ref{optimization} and~\ref{res:strict}.

\begin{table*}[t]
\centering\small
\begin{tabular}{@{}llrrrrr@{}}
\toprule
Optimizer & Objective & BLiMP & Supp. & EWoK & Ent.\ trk. & COMPS \\
\midrule
AdamW & NTP & 78.08 & 58.88 & 54.89 & 18.85 & 56.82 \\
AdamW & NextLat & 77.95 & 59.92 & 55.07 & 19.63 & 57.69 \\
Muon/LambW & NTP & 79.81 & 60.03 & 56.66 & 20.84 & 57.84 \\
Muon/LambW & NextLat & 79.75 & 61.01 & 57.17 & 20.27 & 57.90 \\
\bottomrule
\end{tabular}
\caption{Optimizer-recipe comparison on the strict track: 12-layer, 768-dimensional DAT models (9SA/3RA, SwiGLU, relative symbols), batch of 64 sequences. AdamW: lr $5\times10^{-4}$, no weight decay. Muon/LambW: Muon lr 0.02 (momentum 0.95, 5 Newton--Schulz steps) for hidden matrix parameters; LambW lr $10^{-3}$ (weight decay 0.01) for embeddings, heads, norms, and biases. All rows are seed 0. NTP models use tied LM heads; NextLat models use untied LM heads. The Muon/LambW recipe improves all five measures, but the comparison jointly changes optimizer, learning rate, and weight decay.}
\label{tab:recipe_comparison}
\end{table*}

\paragraph{Strict-track fine-tuned task performance.}
Table~\ref{tab:finetune_results} reports the fine-tuned (Super)GLUE comparison between the same two 12-layer Muon/LambW models as Table~\ref{tab:recipe_comparison}, discussed in Section~\ref{res:nextlat}.

\begin{table}[t]
\centering\small
\begin{tabular}{@{}lrrr@{}}
\toprule
Task & Muon base & Muon NL & $\Delta$ \\
\midrule
BoolQ  & 0.684 & 0.684 & $0.000$ \\
MNLI   & 0.596 & 0.594 & $-0.002$ \\
MRPC   & 0.725 & 0.740 & $+0.015$ \\
MultiRC & 0.598 & 0.655 & $+0.057$ \\
QQP    & 0.761 & 0.772 & $+0.011$ \\
RTE    & 0.547 & 0.576 & $+0.029$ \\
WSC    & 0.654 & 0.692 & $+0.038$ \\
\bottomrule
\end{tabular}
\caption{Fine-tuned (Super)GLUE task accuracies: Muon/LambW base (no NextLat, tied head) vs.\ Muon/LambW NextLat (untied head). All seven tasks are scored by accuracy (MultiRC by per-answer accuracy, rather than SuperGLUE's F1a and exact match); the leaderboard itself scores MRPC and QQP by F1, so these values differ from the corresponding leaderboard entries for the same checkpoints. NextLat improves 5 of 7 task accuracies; a two-sided paired $t$-test across the seven task accuracies gives $t(6)=2.61$, $p=0.040$. These two strict-track models differ in LM-head tying and parameter count ($\approx 12.6$M); the strict-small matrix separates the objective from tying, since its untied/NTP and untied/NextLat cells are matched on both.}
\label{tab:finetune_results}
\end{table}

\paragraph{Reading regression.}
Table~\ref{tab:reading_regression} gives the fixed-effect coefficients of the cross-experiment reading regression of Sections~\ref{stats} and~\ref{res:nextlat}.

\begin{table}[t]
\centering\small

\resizebox{\columnwidth}{!}{%
\begin{tabular}{@{}lrrrr@{}}
\toprule
Term & Estimate & SE & $t$ & $p$ \\
\midrule
(Intercept)         & $\phantom{-}0.048$  & $0.011$ & $\phantom{-}4.56$  & $0.001$ \\
model\_type         & $-0.001$ & $0.000$ & $-2.61$ & $0.010$ \\
objective           & $\phantom{-}0.002$  & $0.000$ & $\phantom{-}5.22$  & $<0.001$ \\
tie\_lm\_head       & $\phantom{-}0.000$  & $0.000$ & $\phantom{-}0.35$  & $0.728$ \\
datapoint\_length1 & $-0.005$ & $0.001$ & $-7.08$ & $<0.001$ \\
datapoint\_length2 & $-0.004$ & $0.001$ & $-4.85$ & $<0.001$ \\
model\_variant      & $-0.003$ & $0.000$ & $-28.48$ & $<0.001$ \\
\bottomrule
\end{tabular}%
}
\caption{Cross-experiment reading regression: coefficients for the incremental $R^2$ ($\Delta R^2$) gained by adding model surprisal to baseline reading-time predictors. The response is the per-experiment, per-dependent-variable $\Delta R^2$. The training objective is significant ($t=5.22$, $p<0.001$): Next-Latent Prediction models explain more variance in human reading measures than NTP models. Model type (DAT vs.\ standard) is also significant ($p=0.010$). The model-variant effect ($p<0.001$) reflects the large difference between base and spillover reading conditions.}
\label{tab:reading_regression}
\end{table}

\paragraph{Architecture-comparison group means.}
Table~\ref{tab:main_results} gives the per-configuration means behind Section~\ref{res:arch}.

\begin{table*}[t]
\centering\small

\begin{tabular}{@{}llccrrrr@{}}
\toprule
Architecture & Objective & LM head & $N$ & BLiMP & Reading & Entity & COMPS \\
\midrule
Standard Transformer & NTP     & tied   & 3 & 67.76 & 7.06 & 14.27 & 52.10 \\
Standard Transformer & NTP     & untied & 3 & 68.16 & 7.16 & 19.14 & 51.86 \\
Standard Transformer & NextLat & untied & 3 & 67.68 & 7.36 & 16.64 & 52.03 \\
\addlinespace
  DAT (RCA, SwiGLU)    & NTP     & tied   & 3 & \textbf{70.66} & 6.66 & 19.71 & 52.42 \\
  DAT (RCA, SwiGLU)    & NextLat & untied & 3 & 70.62 & 7.27 & \textbf{21.92} & 52.36 \\
  DAT (RA, GELU)       & NTP     & tied   & 3 & 70.04 & 6.89 & 22.23 & \textbf{52.51} \\
  DAT (RA, GELU)       & NextLat & untied & 3 & 69.39 & \textbf{7.37} & 20.46 & 52.28 \\
\bottomrule
\end{tabular}
\caption{Main results: BLiMP accuracy and downstream benchmarks by architecture and training objective. Standard Transformer baselines and DAT models share the same 12-layer, 768-dimensional, 12-head configuration, but also differ in feedforward activation (GELU vs.\ SwiGLU), initialization, and symbol mechanism. $N$ is the number of random seeds. Reading is the composite reading score (mean of eye-tracking and self-paced reading).}
\label{tab:main_results}
\end{table*}

\begin{figure}[t]
  \centering
  \includegraphics[width=0.95\columnwidth]{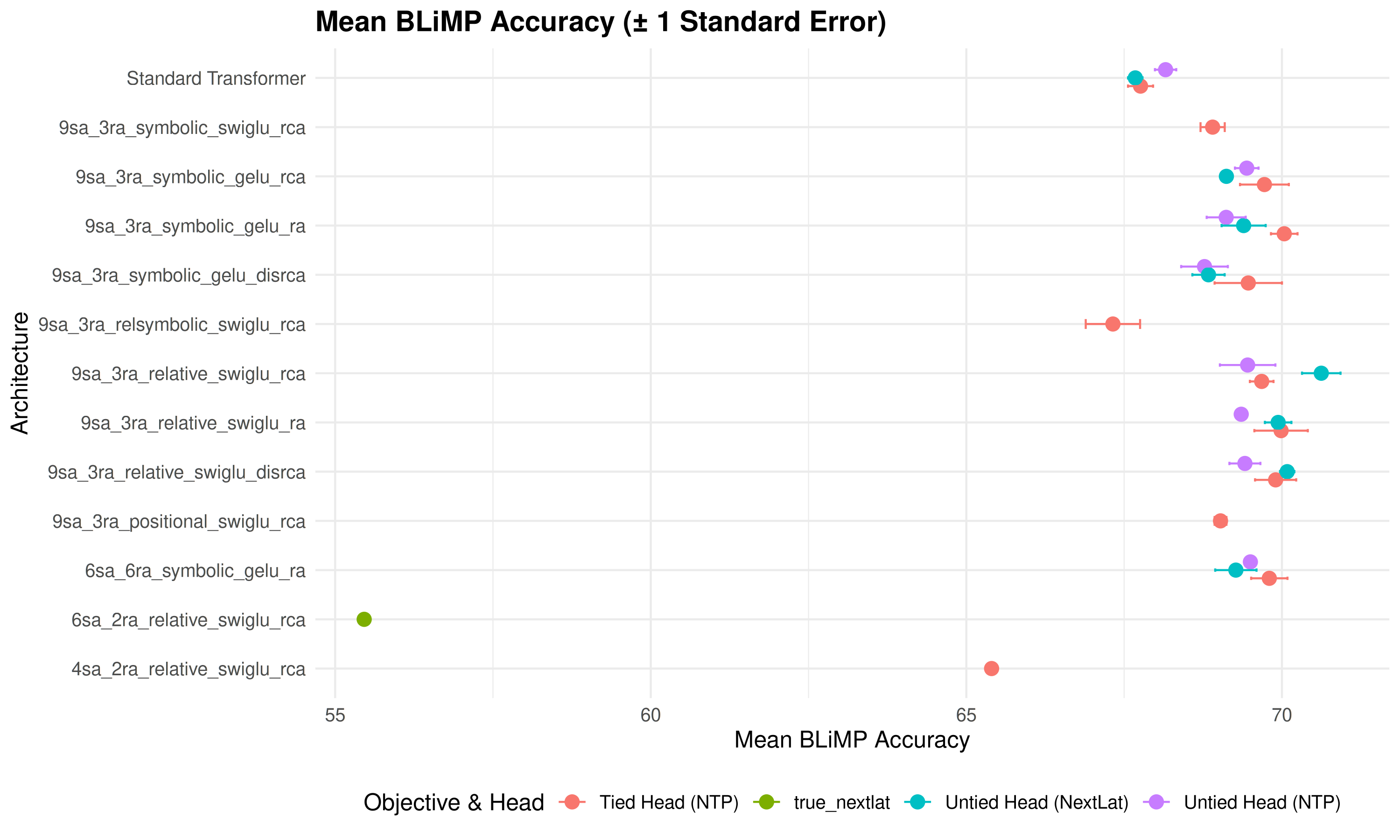}
  \caption{Mean BLiMP accuracy across model configurations. With few exceptions, DAT architectures outperform the standard Transformer baseline (top). Error bars are $\pm 1$ standard error across seeds.}
  \label{fig:blimp_means}
\end{figure}

\paragraph{Symbol ablation at submission scale (not controlled).}
A strict-small run with a learned relative-symbol library in place of RoPE-based symbols scores 71.42 on BLiMP against 72.33 for the submitted model, but is 14 layers deep rather than 12 and was trained with an exponential moving average, so the difference is not attributable to the symbol mechanism. A depth-matched 12-layer run completed training but was not evaluated.

\paragraph{Attention type at 100M words (18 layers).}
Two 18-layer curriculum runs differ only in attention type and in batch size (48 for RCA, 44 for RA; the RA model's relation projections raise its memory cost): same 2026 tokenizer, same phase-A switch point, same learning rates, weight decay, and epoch counts, each warm-started from its own phase~A. RA scores higher on six of seven benchmarks --- BLiMP 79.30 vs.\ 78.44, EWoK 57.93 vs.\ 56.20, entity tracking 22.77 vs.\ 19.07, COMPS 58.62 vs.\ 57.23, and both GlobalPIQA splits --- and lower on the BLiMP supplement (68.56 vs.\ 69.00). Each arm is a single seed, and RA is sensitive to the surrounding hyperparameters: two further 18-layer RA runs under different learning-rate, switch-point, and epoch settings score 78.18 and 76.49, a range that spans the RCA result. All figures in this paragraph are raw-checkpoint scores, not exponential moving averages.

\paragraph{Wide-12L training stability (controlled for RA vs.\ RCA).}
At 1,024 dimensions and 12 layers, a pair differing only in attention type shows RA diverging during training (loss peaking at 5.25 in epoch 3, never recovering below its epoch-0 value; BLiMP 58.21, EWoK 50.15, COMPS 49.65, all at or near chance) while RCA trains monotonically (BLiMP 79.72). A DisRCA run also trains cleanly (BLiMP 79.38) but additionally differs in weight decay (0.01 vs.\ 0.05). The submitted 16-layer wide RA model trained normally under a two-phase curriculum with a reduced Phase-B learning rate. Per-epoch training loss: RA 3.44, 3.69, 4.55, 5.25, 5.14, 4.97, 4.49, 3.98, 3.75, 3.61; RCA 3.34, 2.81, 2.67, 2.40, 2.36, 2.33, declining monotonically; DisRCA 3.36, 2.95, 2.90, 2.87, 2.84, 2.82, 2.78, 2.74, 2.70, 2.68.

\paragraph{Tokenizer and curriculum observations (not controlled).}
Two strict-track observations from the 18-layer curriculum experiments are worth recording, though neither is a controlled comparison. First, the 2024-tokenizer run scores slightly higher than the 2026 retraining on the same 18-layer architecture (BLiMP EMA 79.17 vs.\ 78.44; supplement 70.45 vs.\ 69.00), but the runs also differ in weight decay (0.01 vs.\ 0.05), phase-A switch point, and Phase-B epoch count, so the tokenizer effect cannot be isolated. Second, two 2026-tokenizer RA curriculum runs with different hyperparameter settings (learning rate 4e-4 vs.\ 1e-3, Muon lr 0.008 vs.\ 0.02, phase-A switch at checkpoint 3 vs.\ 1, Phase-B epochs 6 vs.\ 8) score BLiMP 79.30 and 76.49; the lower-learning-rate, later-switch setting wins. A curriculum-versus-document-boundary pair is confounded by warm-start, learning rate, Muon lr, and epoch count (weight decay is identical).

\paragraph{Baseline-vs-DAT omnibus tests.}
Table~\ref{tab:anova_baseline_vs_dat} reports the Type~III Wald $\chi^2$ tests for the baseline-vs-DAT GLMM of Section~\ref{res:arch}, and Table~\ref{tab:contrasts_rca_symbols} the full pairwise contrasts for the symbol-retrieval comparison of Section~\ref{res:rca}.

\begin{table}[t]
\centering\small

\begin{tabular}{@{}lrrr@{}}
\toprule
Effect & $\chi^2$ & df & $p$ \\
\midrule
(Intercept)          & 39.76  & 1 & $<0.001$ \\
base\_arch           & 139.43 & 1 & $<0.001$ \\
is\_tied             & 2.64   & 1 & 0.105 \\
objective            & 2.26   & 1 & 0.132 \\
base\_arch:is\_tied  & 10.62  & 1 & 0.001 \\
base\_arch:objective & 10.81  & 1 & 0.001 \\
\bottomrule
\end{tabular}
\caption{Type III ANOVA (Wald $\chi^2$ tests) for the baseline-vs-DAT BLiMP model (binomial GLMM). The model includes fixed effects for base architecture (standard Transformer vs.\ DAT), LM-head tying, and training objective, with crossed random intercepts for experiment, BLiMP subtest, and pseudo-item. The base-architecture effect is large and significant; the architecture also interacts significantly with both head tying and objective.}
\label{tab:anova_baseline_vs_dat}
\end{table}

\begin{table}[t]
\centering\small

\resizebox{\columnwidth}{!}{%
\begin{tabular}{@{}llrrr@{}}
\toprule
Contrast & Estimate & SE & $z$ & $p$ \\
\midrule
relative $-$ relative\_rope       & $-0.006$ & $0.019$ & $-0.30$ & $0.999$ \\
relative $-$ positional           & $\phantom{-}0.012$ & $0.019$ & $\phantom{-}0.61$ & $0.997$ \\
relative $-$ positional\_sinusoidal & $\phantom{-}0.030$ & $0.019$ & $\phantom{-}1.54$ & $0.722$ \\
relative $-$ symbolic             & $\phantom{-}0.028$ & $0.019$ & $\phantom{-}1.42$ & $0.792$ \\
relative $-$ relsymbolic          & $\phantom{-}0.062$ & $0.019$ & $\phantom{-}3.21$ & $0.023$ \\
relative $-$ relsymbolic\_n4      & $\phantom{-}0.182$ & $0.019$ & $\phantom{-}9.37$ & $<0.001$ \\
\bottomrule
\end{tabular}%
}
\caption{RCA/SwiGLU symbol-retrieval contrasts from the BLiMP subtest binomial GLMM. Contrasts are relative to the \texttt{relative} (learned relative symbol) baseline; $p$-values are Tukey-adjusted.}
\label{tab:contrasts_rca_symbols}
\end{table}

\paragraph{Internal-DAT analysis.}
The internal-DAT ANOVA (Table~\ref{tab:anova_internal_dat}) decomposes BLiMP performance across the DAT design factors. The activation/symbol-position term ($\chi^2(2)=26.41$, $p<0.001$) covers the early GELU/symbolic runs that preceded the fixed-SwiGLU design (Section~\ref{experiments-design}); LM-head tying ($p=0.006$) and context length ($p=0.001$) are also significant. Relational attention type ($p=0.174$) and layer allocation ($p=0.437$) are not (Section~\ref{res:ratype}). The training objective is also not significant within the DAT family on BLiMP ($p=0.216$), consistent with architecture being the primary lever for structural generalization while NextLat provides a complementary cognitive-alignment benefit.

\begin{table}[t]
\centering\small

\begin{tabular}{@{}lrrr@{}}
\toprule
Effect & $\chi^2$ & df & $p$ \\
\midrule
(Intercept)           & 56.999 & 1 & $<0.001$ \\
attn\_type            & 3.496  & 2 & 0.174 \\
layers                & 0.604  & 1 & 0.437 \\
activation\_position  & 26.406 & 2 & $<0.001$ \\
is\_tied              & 7.564  & 1 & 0.006 \\
objective             & 1.531  & 1 & 0.216 \\
context\_length       & 10.187 & 1 & 0.001 \\
\bottomrule
\end{tabular}
\caption{Type III ANOVA (Wald $\chi^2$ tests) for internal DAT mechanics on BLiMP (binomial GLMM). Fixed effects: relational attention type (RA/RCA/DisRCA), layer allocation, activation/position setting, LM-head tying, training objective, and context length. Activation/position, head tying, and context length are significant; relational attention type and layer allocation are not.}
\label{tab:anova_internal_dat}
\end{table}

\paragraph{Objective at matched LM-head tying.}
The architecture comparison observes three of the four head-tying $\times$ objective combinations: tied/NTP, untied/NTP, and untied/NextLat. The last two are matched on head tying and on parameter count, since the auxiliary dynamics model is discarded after training, so their difference isolates the objective. Within the controlled cell of Table~\ref{tab:ra_type_cells}, NextLat gains 0.83 BLiMP points over NTP (RCA $+1.17$, DisRCA $+0.67$, RA $+0.58$); the internal-DAT model does not find the objective significant on BLiMP ($p=0.216$), while it is highly significant in the reading regression ($t=5.22$, $p<0.001$). The first two cells isolate LM-head tying at NTP: 70.18 tied against 69.42 untied.

\paragraph{RA-type per-cell means.}
Table~\ref{tab:ra_type_cells} reports BLiMP means for the controlled RA-type comparison of Section~\ref{res:ratype}: 25 runs holding SwiGLU, learned relative symbols, 512-token context, 9SA/3RA, identity relation activation, 12 layers, 768 dimensions, and all training hyperparameters fixed. Three of the seven RA runs used a batch of 16; the other 22 used 12. Pooled within-cell seed SD is 0.54 (df=16).

\begin{table}[t]
\centering\footnotesize
\setlength{\tabcolsep}{3pt}
\begin{tabular}{@{}lrrrr@{}}
\toprule
Condition & RA & RCA & DisRCA & spread \\
\midrule
tied / NTP       & 69.99 (3) & 70.66 (3) & 69.90 (3) & 0.76 \\
untied / NTP     & 69.36 (2) & 69.46 (3) & 69.41 (3) & 0.10 \\
untied / NextLat & 69.94 (2) & 70.62 (3) & 70.08 (3) & 0.68 \\
\midrule
marginal         & 69.79 (7) & 70.25 (9) & 69.80 (9) & 0.45 \\
\bottomrule
\end{tabular}
\caption{BLiMP cell means for the controlled RA-type comparison. RCA is numerically highest in all three conditions; the omnibus GLMM finds no significant effect of attention type ($\chi^2(2)=3.50$, $p=0.174$).}
\label{tab:ra_type_cells}
\end{table}

\paragraph{Further symbol-retrieval results.}

Table~\ref{tab:rca_symbols} gives the per-condition means behind Section~\ref{res:rca}, and Table~\ref{tab:rca_symbols_reading} the same comparison on composite reading scores.

\begin{table}[t]
\centering\small

\begin{tabular}{@{}lccccc@{}}
\toprule
Symbol condition & BLiMP & Entity & COMPS\\
\midrule
relative (learned)         & 69.34 & 22.13 & 52.56 \\
  relative\_rope             & \textbf{69.43} & 20.54 & 52.54 \\
positional                 & 69.17 & 16.69 & 52.56 \\
  positional\_sinusoidal     & 68.88 & \textbf{23.12} & 52.14 \\
  symbolic                   & 68.90 & 17.70 & \textbf{52.60} \\
relsymbolic                & 68.33 & 21.17 & 52.09 \\
relsymbolic\_n4            & 66.32 & 19.88 & 51.63 \\
\bottomrule
\end{tabular}
\caption{RCA/SwiGLU symbol-retrieval comparison, using 5 seeds on BLiMP, COMPS and Entity accuracy scores. All models use the same 9SA/3RA RCA architecture with RoPE positional encodings and SwiGLU activations. Further comparisons are in the appendix.}
\label{tab:rca_symbols}
\end{table}

\begin{table}[t]
\centering\small

\begin{tabular}{@{}lcccc@{}}
\toprule
Symbol condition & Eye-tracking & SPR \\
\midrule
relative (learned)         & 9.94 & 3.64 \\
relative\_rope             & 9.41 & 3.66 \\
positional                 & 9.78 & 3.81 \\
positional\_sinusoidal     & 9.84 & 3.70 \\
symbolic                   & 9.64 & 3.58 \\
relsymbolic                & 9.93 & 3.85 \\
relsymbolic\_n4            & 10.06 & 3.90 \\
\bottomrule
\end{tabular}
\caption{RCA/SwiGLU symbol-retrieval comparison (complete design: 7 symbol conditions $\times$ 5 seeds). All models use the same 9SA/3RA RCA architecture with RoPE positional encodings and SwiGLU activations. The \texttt{relative} condition is the reference baseline for contrasts. Shown measures are the eye-tracking and self-paced reading composite scores.}
\label{tab:rca_symbols_reading}
\end{table}

\begin{table}[t]
\centering\small

\begin{tabular}{@{}lrrr@{}}
\toprule
Effect & $\chi^2$ & df & $p$ \\
\midrule
(Intercept)        & 38.28  & 1  & $<0.001$ \\
symbol\_retrieval  & 134.29 & 6  & $<0.001$ \\
\bottomrule
\end{tabular}
\caption{Type III ANOVA (Wald $\chi^2$ tests) for the RCA/SwiGLU symbol-retrieval BLiMP subtest model (binomial GLMM). The symbol-retrieval mechanism has a large, significant effect on BLiMP subtest accuracy ($\chi^2(6)=134.29$, $p<0.001$).}
\label{tab:anova_rca_symbols}
\end{table}

\begin{table}[t]
\centering\small
\begin{tabular}{@{}rrrrrr@{}}
\toprule
SA & RA & BLiMP & Supp. & EWoK & COMPS \\
\midrule
1  & 11 & 68.22 & 56.92 & 50.01 & 51.67 \\
3  & 9  & 68.22 & 57.38 & 50.37 & 51.70 \\
4  & 8  & 69.23 & 56.92 & 50.44 & 51.77$^{\dagger}$ \\
6  & 6  & \textbf{70.33} & 56.32 & 50.55 & 51.64 \\
8  & 4  & 69.18 & 57.06 & 50.75 & 51.70 \\
9  & 3  & 69.70 & \textbf{57.75} & 50.60 & \textbf{52.04} \\
11 & 1  & 69.67 & 56.87 & \textbf{51.13} & 51.81 \\
\bottomrule
\end{tabular}
\caption{RCA SA/RA head-ratio sweep at 512-token context, mean of 3 seeds. All models use RCA with SwiGLU and relative symbols. The 6SA/6RA split achieves the best BLiMP accuracy (70.33\%). $^{\dagger}$One 4SA/8RA COMPS evaluation is missing; its COMPS mean is over two seeds.}
\label{tab:sa_ra_ratio}
\end{table}

\begin{figure}[t]
  \centering
  \includegraphics[width=0.95\columnwidth]{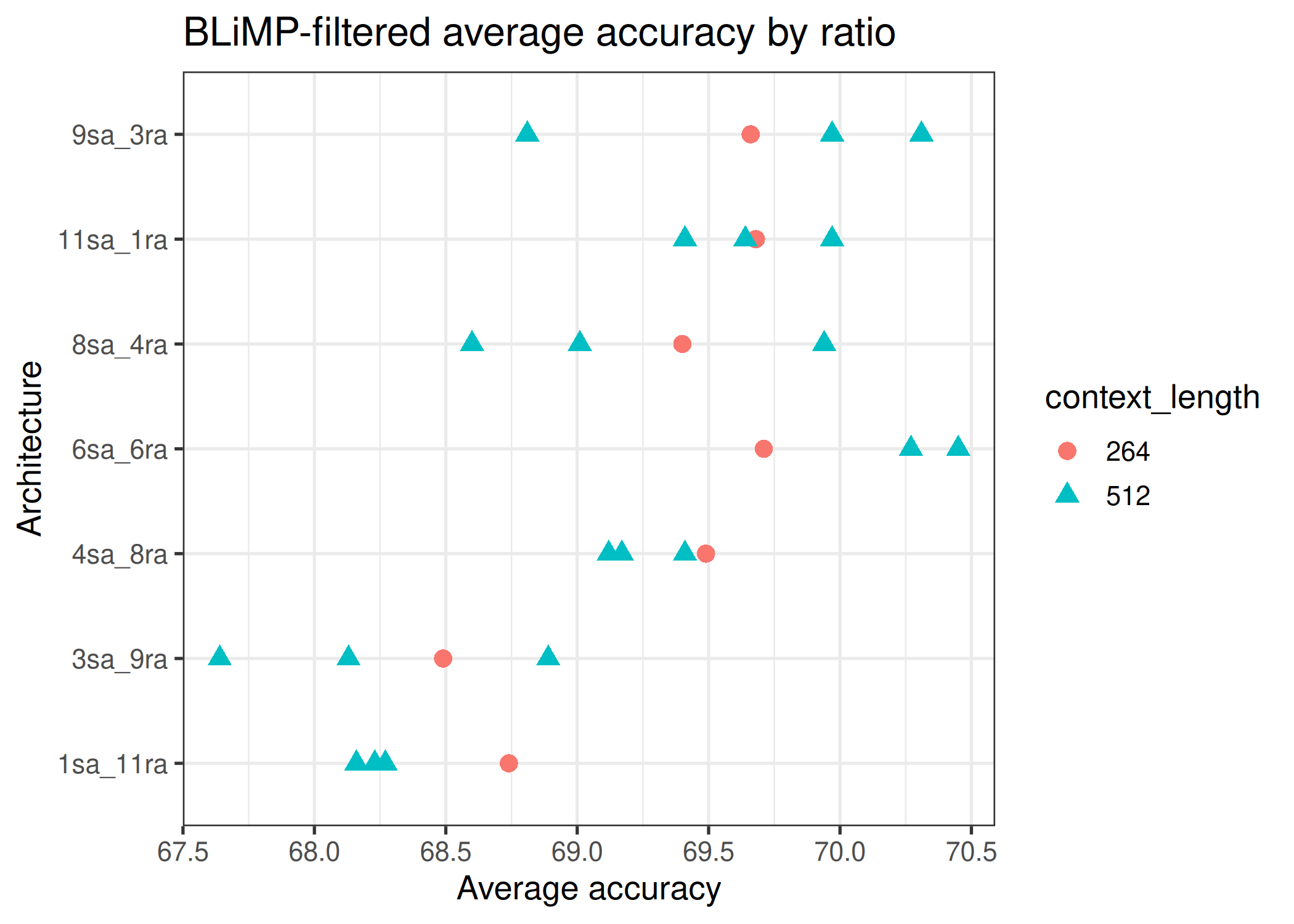}
  \caption{Mean BLiMP accuracy across SA/RA head-ratio splits. Error bars are $\pm 1$ standard error across 3 seeds.}
  \label{fig:sa_ra_blimp}
\end{figure}

\paragraph{SA/RA head-ratio sweep.}
Figure~\ref{fig:sa_ra_blimp} shows mean BLiMP accuracy across SA/RA head-ratio splits, and Table~\ref{tab:ra_sa_ratio_lmer} reports the mixed-effects model estimates for the SA/RA head-ratio sweep at 512-token context (3 seeds per ratio, seed as random intercept). The RA head fraction has a significant negative effect on BLiMP and EWoK, but no significant effect on COMPS or reading measures.

\begin{table}[t]
\centering\footnotesize
\setlength{\tabcolsep}{3pt}
\begin{tabular}{lrrrr}
\toprule
Metric & Estimate (Slope) & Std. Error & $t$ & $p$ \\
\midrule
BLiMP & -1.827 & 0.310 & -5.887 & <0.001*** \\
Reading & 0.170 & 0.240 & 0.710 & 0.488 \\
COMPS & -0.245 & 0.190 & -1.287 & 0.201 \\
EWoK & -1.039 & 0.447 & -2.323 & 0.021* \\
\bottomrule
\end{tabular}

\caption{Mixed-effects models for the SA/RA head-ratio sweep at 512-token context (3 seeds per ratio, seed as random intercept). The RA head fraction has a significant negative effect on BLiMP and EWoK, but no significant effect on COMPS or reading measures. The seed random effect collapsed to zero variance in the BLiMP model.}
\label{tab:ra_sa_ratio_lmer}
\end{table}

\end{document}